\documentclass[conference,letterpaper]{IEEEtran}

\usepackage[utf8]{inputenc} 
\usepackage[T1]{fontenc}
\usepackage{url}
\usepackage{ifthen}
\usepackage[numbers,sort&compress]{natbib}
\usepackage{wrapfig}
\usepackage{amssymb}
\usepackage{graphicx}
\usepackage[cmex10]{amsmath} % Use the [cmex10] option to ensure complicance
\begin{document}
\title{GQ-VAE: A gated quantized VAE \\for learning variable length tokens} 

% %%% Single author, or several authors with same affiliation:
% \author{%
%  \IEEEauthorblockN{Author 1 and Author 2}
% \IEEEauthorblockA{Department of Statistics and Data Science\\
%                    University 1\\
 %                   City 1\\
  %                  Email: author1@university1.edu}% }

%%% Several authors with up to three affiliations:

\author{%
  \IEEEauthorblockN{Theo Datta, Kayla Huang, Sham Kakade, David Brandfonbrener}
  \IEEEauthorblockA{Kempner Institute, Harvard University\\
                    All correspondance to:  theodatta@college.harvard.edu}
  % \and
  % \IEEEauthorblockN{Author 2 and Author 3}
  % \IEEEauthorblockA{Research Center XY\\ 
  %                   City 2\\
  %                   Email: \{author2, author3\}@research-center.com}
}

\maketitle

%%%%%%
%% Abstract: 
%% If your paper is eligible for the student paper award, please add
%% the comment "THIS PAPER IS ELIGIBLE FOR THE STUDENT PAPER
%% AWARD." as a first line in the abstract. 
%% For the final version of the accepted paper, please do not forget
%% to remove this comment!
%%

\begin{abstract}
   While most frontier models still use deterministic frequency-based tokenization algorithms such as byte-pair encoding (BPE), there has been significant recent work to design learned neural tokenizers. However, these schemes generally add to underlying language model complexity and force large changes to architecture, making them hard to implement at large scales. To overcome these challenges, we propose the gated quantized variational autoencoder (GQ-VAE), a novel architecture that can be independently pre-trained to serve as a drop-in replacement for existing tokenizers. The key innovation of the architecture is to learn to encode variable-length discrete tokens. GQ-VAE improves compression and language modeling performance over a standard VQ-VAE tokenizer, and approaches the compression rate and language modeling performance of BPE. Interestingly, if we use BPE with a smaller vocabulary, such that the compression is equivalent between GQ-VAE and BPE, we find that GQ-VAE improves downstream language model learning. We conclude with a discussion of several exciting avenues for future work. Code can be found at https://github.com/Theo-Datta-115/gq-vae. 
\end{abstract}

\section{Introduction}

Tokenization is the first step of any model language modeling pipeline, and it plays a crucial role in encoding semantic information and compressing input text. Yet despite recent progress in language modeling, tokenization has remained relatively unchanged, reliant on traditional compression algorithms like BPE \cite{sennrich2015neural, touvron2023llama, achiam2023gpt}. BPE builds a fixed mapping of tokens to indexes by iteratively merging the most frequent pairs of symbols to form a compact vocabulary. This method is imperfect, as research has shown that language models could potentially unlock large gains by improving the ability of tokenizers to compress data and extract learnable subword chunks. \cite{bostrom_byte_2020, schmidt_tokenization_2024, wang_tokenization_2024, goldman_unpacking_2024}.

One obvious direction to explore in solving this challenge is to engineer tokenization schemes that are learned dynamically instead of deterministically. There have been numerous attempts to create end-to-end Transformers that use a dynamic encoder-decoder structure to learn to 'tokenize' blocks of characters or bytes into continuous representations. \cite{clark_canine_2022, xue_byt5_2022, tay_charformer_2022, yu_megabyte_2023, ho_block_2024, pagnoni_byte_nodate, neitemeier_hierarchical_2025}. However, this approach comes with drawbacks. The block-based structure of these models often fails to encode variable length tokens, an important semantic feature of language. They also force the underlying transformers to learn a far more complex compression structure than BPE's discrete vocabulary mapping, potentially restricting scalability. Finally, they require changing Large Language Model (LLM) architectures to include large encoder-decoder layers, reducing the applicability of years of research into optimal training dynamics. 

\begin{figure}[htbp]
  \centering
  \includegraphics[width=0.4\textwidth]{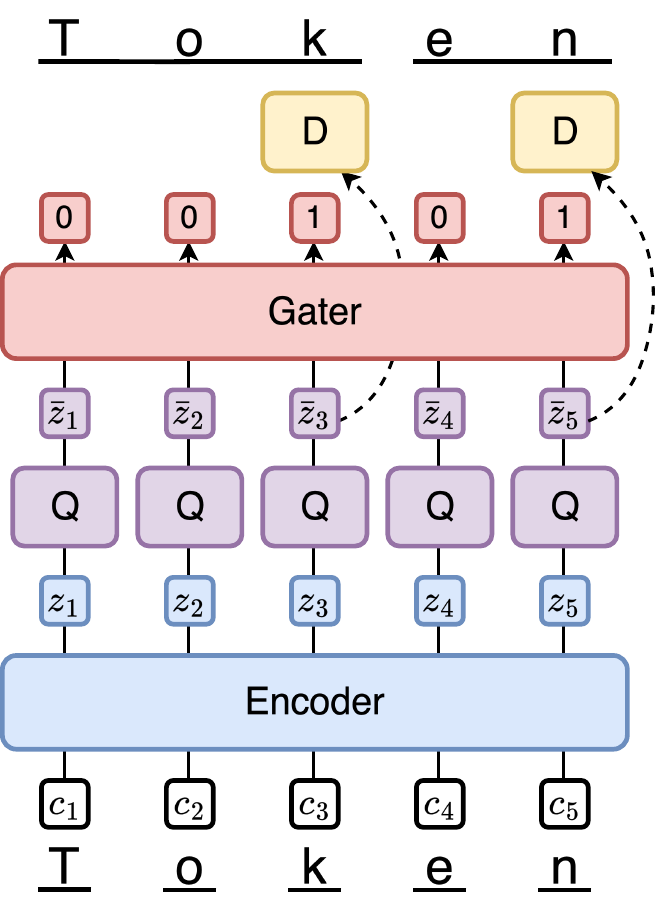}
  \caption{GQ-VAE Architecture. D=Decoder, Q=Quantizer. Encoder and Gater are transformers. The decoder head is illustrated in (Figure \ref{fig:decoder}).}
  \label{fig:arch}
\end{figure}

To overcome these challenges, we introduce GQ-VAE (Figure \ref{fig:arch}), a standalone model that learns discrete, variable-length tokenization and can output a Hugging Face-compatible tokenizer. While we lag slightly behind the compression level of BPE, our model is able to beat fixed-chunk baselines and generates text that is more learnable for LLMs. We hope this proof-of-concept model is instructive for future attempts to model discrete tokenization processes.

\section{GQ-VAE Architecture }\label{sec:method}

\subsection{Architecture and inference procedure}

We aim to model discrete, variable-length toknization as seen in language encoding tasks. This process comes with several key challenges. To model discreetness, the model must reduce inputs of bytes or characters to some discrete latent codebook. To model compression, it must learn to represent the input in a reduced number of these tokens. Finally, the decoder must be able to differentiate both the content of these tokens and their variable length. All these features must be tied together by a loss function that is sufficiently continuous and convex to be learnable.

\begin{figure}[htbp]
  \centering
  \includegraphics[width=0.3\textwidth]{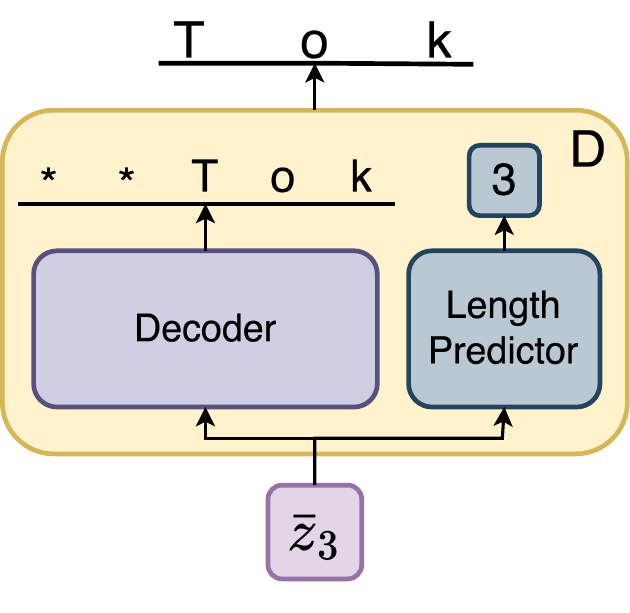}
  \caption{Decoder Head}
  \label{fig:decoder}
\end{figure}

To accomplish this task our model has four parts: an encoder $ E $ that maps characters/bytes to latent vectors, a quantizer $ Q$ that quantizes the latent vectors to a discrete dictionary, a gater $ G$ that gates the quantized vectors to define token boundaries, and a decoder $ D $ that decodes a quantized latent vector to a variable number of characters $w$, parameterized as a fixed number along with a length $l$ that determines which characters to mask.

More formally, if we are given a sequence of characters $ c_1, \dots, c_T$ our model works as follows:
The encoder outputs $ z_t$ and $g_t$ for each $t$ given the entire input sequence, as seen in Figure \ref{fig:arch}. 
\begin{align}
    &z_t = E(c_0, \dots, c_T)_t &z_t \in \mathbb{R}^d\\
    &\bar z_t = Q(z_t)  &\bar z_t \in \{z^{(1)}, \dots, z^{(|V|)}\}\\
    &g_t = G(\bar z_0, \dots, \bar z_T)_i &  g_t \in [0,1]\\
    &\hat c_t^{w-1}, \dots \hat c_t^0, \hat l_t =  D(\bar z_t) &\hat c_t^i \in \mathbb{R}^{|C|},\ \hat l_t \in \mathbb{R}^{w}
\end{align}

At inference, we use the gates $g_t$ to pass through only latent vectors where $g_t > 1/2$. As a result, our model can compress the input into a smaller subset of latent codebook vectors. Additionally, since $g_t$ is learned and flexible, our model can encode tokens of variable length, and the length prediction head allows us to decode these lengths consistently.

\subsection{Training}

We train an autoencoder with two main goals: reconstruct the input characters and compress the input by using fewer tokens. We use a transformer as an encoder and a transposed convolutional layer as the main decoder to project tokens into sequences of length $w$, representing the layers "Encoder" and "Decoder" in Fig. \ref{fig:arch} and \ref{fig:decoder}. We use a custom loss function, which is the sum of the following losses. 

\paragraph{Reconstruction.} The reconstruction loss at time $ t$ is defined as follows:
\begin{align}
        \ell_t^{(r)} = \sum_{i=0}^{w-1} m_t^i \cdot \ell( \hat c_{t}^i , c_{t-i})
\end{align}
$$m_t^i = \prod_{j=1}^i (1- g_{t-j}), \qquad m_t^0 = 1$$

We construct a mask $m_t$ for the individual byte losses, as a function of $g_t$, to incentivize the latent tokens to correctly decode a number of bytes based on compression of the gate values. For example, in Figure \ref{fig:arch}, we third gate has value one, and therefore we want the model to correctly decode the first three bytes of the token, but we are agnostic to loss for other bytes encoded in the tokens. 

Therefore the mask value $ m_t^i$ is the probability that character $ t-i$ is decoded from the latent $ \bar z_t$ if we assume that $ g_t = 1$. Note: this loss will try to send all $ g_t \to 1$ since character level reconstruction is easiest. We balance this out by adding a compression loss below. 

\paragraph{Compression.}
We also want to encourage compression. The size of the "codebook" in our quantizer represents the vocabulary in a tradition BPE mapping. If the size of the codebook is $ |V|$, then the number of bits in the token sequence is roughly $ \log |V| \sum_{t=0}^T g_t$.
So, we can add a compression loss as follows:
\begin{align}
    \ell_t^{(cmp)} = g_t
\end{align}

\paragraph{Length decoding.}
To decode variable length tokens, we need to predict the length $ \hat l_t$ of a token that terminates at time $ t$. We will do this by adding a decoder head, as in Figure \ref{fig:decoder}. We want to predict $m_t$, but can't predict the set of $w$ previous $g_t$ necessary to construct it, as that would require predicting changing previous token lengths from the output of a single token. Therefore, we create a proxy for the monotonically decreasing values of $m_t$. We can supervise this prediction with the following loss, with the cumulative sum function being represented as $c(x)$:
\begin{align}
    \hat m_t &= \frac{m_t^*}{max(m_t^*)}, m_t^*=c(e^{(\hat l_t - min(\hat l_t)})) \\ 
    \ell_t^{(l)} &= \text{detach}(g_t) \cdot \ell(\hat m_t, \text{detach}(m_t))
\end{align}
This process predicts the masks of a token ending at $t$, which we can threshold to determine the length of a token as $\sum_{i=0}^{w-1}\hat m_t^i > 0.5$. Note that we weight by $g_t$ to only require mask prediction on used tokens, and detach $g_t$ and $m_t$ so that gradients are only passed back from $ \hat l_t$ and not through the labels. 

\paragraph{Vector quantization.}
Vector quantization using a VQ-VAE requires a few extra loss terms for the codebook, traditionally called the ``codebook loss'' and ``commitment loss,'' as first formulated in Oord et. al. \cite{oord_neural_2018}. "Codebook loss" pushes the codebook vectors toward encodings, while "commitment loss" pushes the encodings toward the codebook vectors, incentivizing a convergence between the two. Given that the quantization is non-differentiable, there is a "straight-through gradient estimation", where gradients are passed through the quantization step from the encoder to the decoder.
\begin{align}
\ell_t^{(cde)} = \lVert sg(\bar z_t) - z_t \rVert^2_{2} \\
\ell_t^{(cmt)} =  \lVert \bar z_t - sg(z_t) \rVert^2_{2}
\end{align}

\paragraph{Putting it together}
With loss weighting parameters $\alpha, \beta, \gamma$, the full loss for a sequence is then:
\begin{align}
    \ell = \sum_t \ell_t^{(r)} + \gamma \cdot \ell_t^{(l)} + \alpha\cdot \ell_t^{(cmp)} + \ell_t^{(cde)} + \beta \cdot \ell_t^{(cmt)}
\end{align}

\subsection{Implementation Details}

We use tinystories \cite{eldan_tinystories_2023}, a restricted vocabulary synthetic dataset, to train our model. To restrict the size of our computational problem and to mirror the training strategies of existing tokenizers, we split this data on a preprocessing regular expression pattern into substrings of maximum size 16. Interestingly, early experimentation found the GPT-2 regex to be the most learnable for our model of the options presented in Dagan et. al. \cite{dagan_getting_2024}, potentially offering one of the first heuristics on the learnability of different regular expressions.

We also implement a few optimizations to improve the quality of learning in our quantization layer, a potential learning bottleneck. These primarily include warm up, caching, and re-sampling schemes to prevent codebook collapse similar to Lancucki et. al.\cite{lancucki_robust_2020}. We generally reconstruct to tokens of maximum length $w = 10$. 

% We also experiment with the VQ-VAE rotation trick to improve gradient pass-through for training stability \cite{fifty_restructuring_2024}.

At inference time we decode characters and lengths for all codebook tokens to produce a dictionary mapping of latent tokens to semantic tokens. When text is provided, we take the set of tokens for that text to be all latent encodings with $g_t > 0.5$, and then decode those tokens greedily from the dictionary mapping.

\section{Results}\label{sec:results}

\subsection{Compression and reconstruction}

\begin{figure}
\centering
\includegraphics[width=0.4\textwidth]{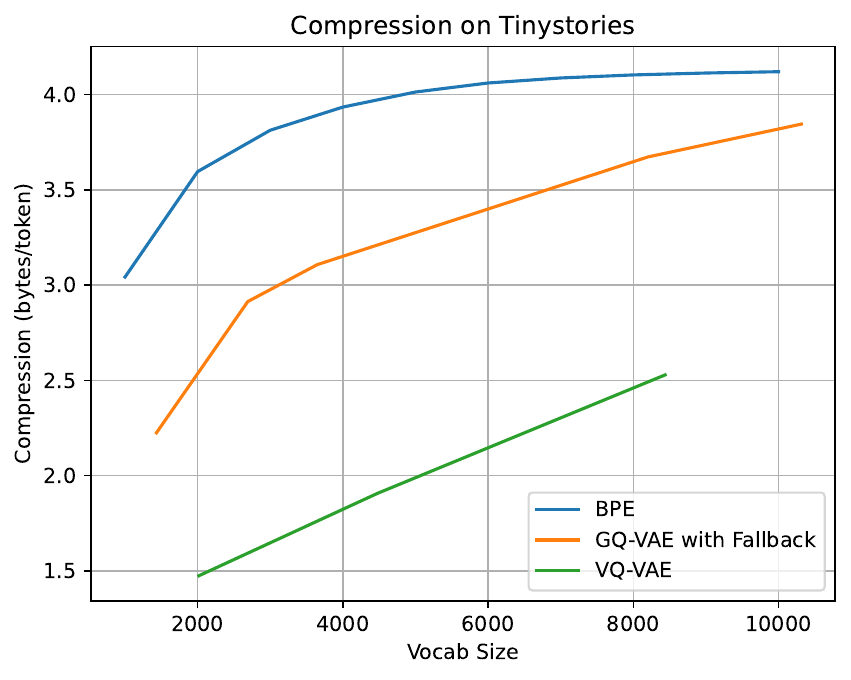}
\caption{Compression and Vocabulary Size. VQ-VAE represents a fixed token length GQ-VAE, and higher vocab results continue the linear trend but are omitted for scale. GQ-VAE vocab sizes are distilled to represent only the number of unique tokens, while the \textit{codebook} sizes in the model are [2k, 5k, 10k, 20k, 50k].}
\label{fig:comp1}
\end{figure}

Our model is able to approach the compression rate of BPE on tinystories (Figure \ref{fig:comp1}). We have two compression metrics for our model: 
\begin{enumerate} \item \textit{Compression without fallback}: this checks the number of gates over 0.5 that the model assigns over a subset of a validation set and divides this by the number of non-padding characters. This is controlled by the hyperparameter $\alpha$, and generally approaches BPE levels at sufficiently large $\alpha$ values. This suggests that further engineering improvements on this model could potentially match or improve on state-of-the-art language compression.

\item \textit{Compression with fallback}: we implement a version of our model with 'fallback', where we check at inference time if input characters are faithfully reconstructed by the model tokenization, and if they are not we replace incorrect tokens with multiple single characters. This ensures that for the GQ-VAE with fallback, all token text is reconstructed without information loss. We see in Figure \ref{fig:comp1} that GQ-VAE can tokenize text with a less than 7\% penalty on compression. \end{enumerate} 

% \begin{figure}
% \centering
% \includegraphics[width=0.4\textwidth]{final_graphs/accuracy.pdf}
% \caption{Accuracy and Vocabulary Size.}
% \label{fig:acc}
% \end{figure}

\begin{figure}
\centering
\includegraphics[width=0.4\textwidth]{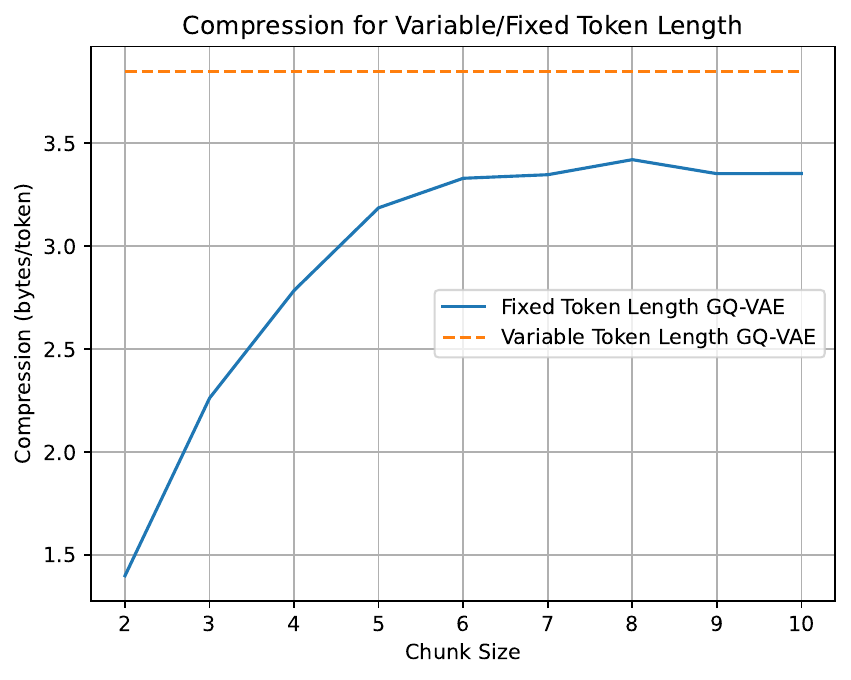}
\caption{Compression for fixed-length token models. These models are all trained on the same hyperparameter setup as the 3.85 bits/byte variable length GQ-VAE baseline.}
\label{fig:hardset}
\end{figure}

To validate that the extra complexity of allowing for variable length tokens is useful, we compare our model to fixed token length baselines. We can modify our GQ-VAE to do fixed-length tokenization by hard-setting $g_t$ to represent some fixed frequency, such as 4-length chunks with $\{0,0,0,1,0,0,0,1\}$. We simply call this baseline VQ-VAE. In Figure \ref{fig:hardset}, we can see that our model achieves higher compression than all fixed token sizes. The flattening in the gains to compressing larger chunk sizees is due to the trade-off in these models between naive compression and the percentage of tokens correctly reconstructed. With fallback, we must replace every token that contains an incorrect character, and therefore every incorrect token reduces compression. Our model achieves higher compression without fallback than a fixed 8-length tokenizer, while reconstructing characters at a comparable rate to the 5-length tokenizer, effectively achieving the best of both worlds. 

 % We hypothesize that some level of incorrect characters is tolerable as there is some amount of inaccuracy (ex. misspellings) within the dataset that our model is correctly ignoring. However, we use fallback to provide results that are maximally comparable with existing tokenization systems. 

% \begin{figure}
% \centering
% \includegraphics[width=0.4\textwidth]{final_graphs/alpha.pdf}
% \caption{Compression and $\alpha$. These runs use the rotation trick in the VQ-VAE to ensure higher training stability.}
% \label{fig:comp2}
% \end{figure}

% Our model also uniquely has a compression hyperparameter, $\alpha$, that can be used to tune compression (Figure \ref{fig:comp2}). While our vocab sweep is run at $\alpha = 2$ to incentivize high compression, we can model lower compression systems on our data by reducing the value of $\alpha$. While current language tasks generally prioritize high compression because of the importance of throughput, this flexibility could be useful for non-language domains.

\subsection{Language Modeling}

\begin{figure}
\centering
\includegraphics[width=0.4\textwidth]{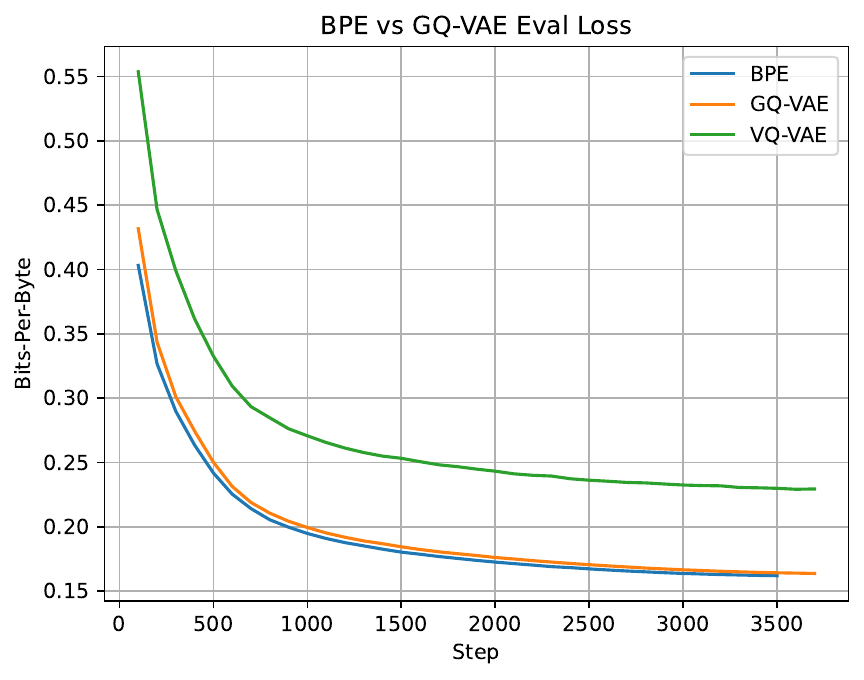}
\caption{Language Modeling with GQ-VAE, BPE, and VQ-VAE (fixed token length GQ-VAE). Models are trained on all of tinystories, so the lower-compression GQ-VAE trains for more steps on the same data. The exact "used" vocabulary sizes for these models are BPE=10000, GQ-VAE=10314, VQ-VAE=13201.}
\label{fig:language2}
\end{figure}

\begin{figure}
\centering
\includegraphics[width=0.4\textwidth]{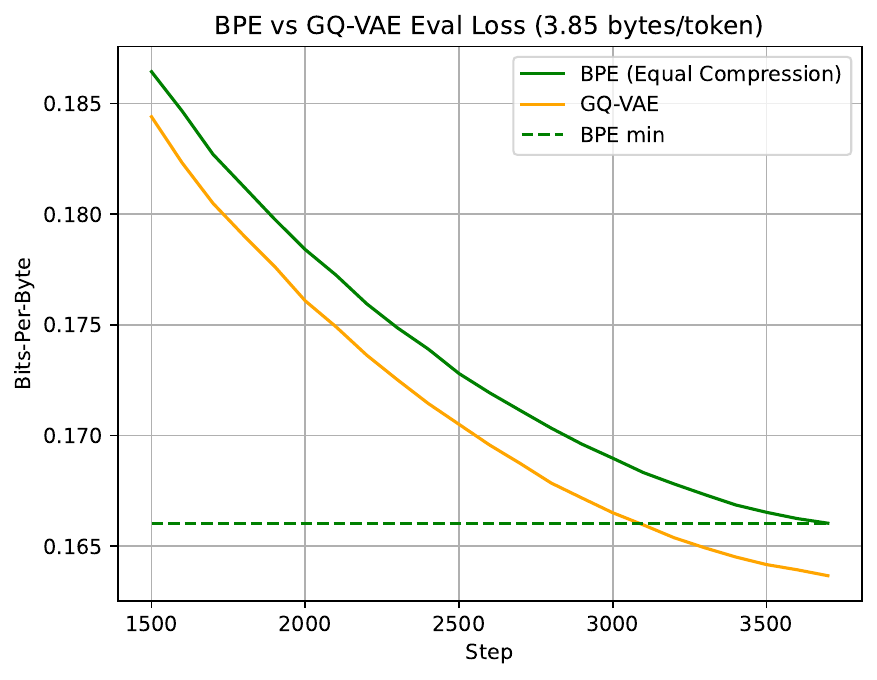}
\caption{Language Modeling with GQ-VAE and BPE, same compression level.}
\label{fig:language}
\end{figure}

One hope of creating a learned tokenizer is that having a gradient descent-based tokenization system would result in a more "learnable" tokenization system that could improve downstream language modeling accuracy. To test this hypothesis, we train an 18 million parameter transformer using the OLMo repository \cite{groeneveld_olmo_2024} on the whole of tinystories, tokenized both by GQ-VAE and BPE. As is visible in Figure \ref{fig:language2}, the lower compression of the GQ-VAE results in it providing 1.07\% higher loss. 

However, we also compare our model to a BPE baseline that has has equivalent compression (3.85 bytes/token) as our chosen GQ-VAE model (Figure \ref{fig:language}), which we achieve by limiting BPE vocabulary size to 3250. We find that the model trained on the GQ-VAE data converges to a lower loss, reaching the BPE loss minimum 600 steps (16.6\%) earlier. This result indicates that the tokenization system created by the GQ-VAE is more learnable for downstream models. 

\begin{figure}
\centering
\includegraphics[width=0.5\textwidth]{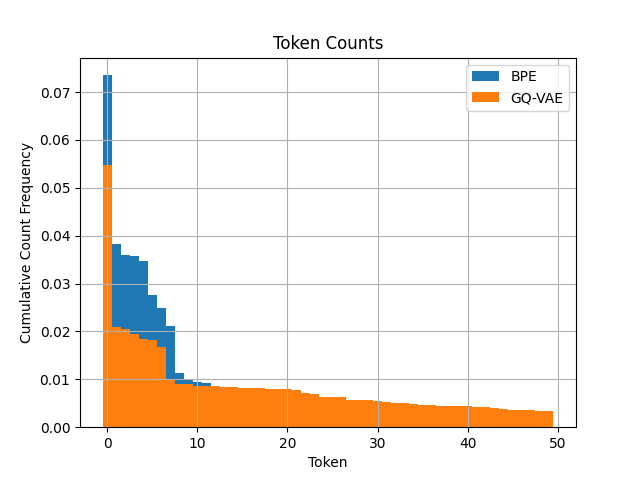}
\caption{Histogram of Token Frequencies. Only the 50 most common tokens are included, but the frequency trend of GQ-VAE having higher frequencies in the distribution's tail continues for later tokens.}
\label{fig:hist}
\end{figure}

To understand what changes in the tokenization system lead to differential language modeling results, we then investigate the distribution of token frequencies under both tokenizers (Figure \ref{fig:hist}). While both systems learn to tokenize common words as complete tokens and therefore approach similar distributions, we observe that the BPE tokenization system results in larger frequencies for the most used tokens. Therefore, GQ-VAE has comparatively more uniform token distribution in the tail of the frequency distribution, something we hypothesize could improve learning over niche tokens. 

\subsection{Limitations}

There are limitations to our work. Our main modeling tests are on tinystories, a toy dataset not used in practical language applications. GQ-VAE runs are oftentimes sensitive to initialization, an artifact of the difficulty of discrete optimization tasks. We think our model is promising, but more engineering challenges must be solved in order to have this model beat existing compression baselines, something we will leave to future work. 

\section{Conclusion}

We present GQ-VAE, a novel proof-of-concept model for discrete, variable-length, independent tokenization, something that to the best of our knowledge has not been attempted in any existing literature. This model demonstrates compression abilities approaching those of state-of-the-art, compression-focused tokenization algorithms, and provides outputs that improve downstream language modeling quality. We believe that this paradigm for tokenization is an interesting and potentially valuable addition to the growing field of novel tokenization methods for language. 

This work opens up new avenues for future exploration. There are clear discrete optimization challenges that must be solved for GQ-VAE compression to surpass BPE. It would also be interesting to see how learned, discrete, variable length compression works on new domains, such as non-English language, or text-free settings like audio and vision. It may also be worth investigating the comparative trade-offs of introducing lossy tokenization algorithms into the language modeling pipeline. We hope that this work contributes to the growing study of learned neural tokenization, a field we think holds great potential for improving the learning and throughput of LLMs.

%%%%%%
%% To balance the columns at the last page of the paper use this
%% command:
%%
%\enlargethispage{-1.2cm} 
%%
%% If the balancing should occur in the middle of the references, use
%% the following trigger:
%%
%\IEEEtriggeratref{7}
%%
%% which triggers a \newpage (i.e., new column) just before the given
%% reference number. Note that you need to adapt this if you modify
%% the paper.  The "triggered" command can be changed if desired:
%%
%\IEEEtriggercmd{\enlargethispage{-20cm}}
%%
%%%%%%

%%%%%%
%% References:
%% We recommend the usage of BibTeX:
%%
\bibliographystyle{IEEEtran}
%\bibliography{definitions,bibliofile}
\bibliography{references}
%%
%% where we here have assumed the existence of the files
%% definitions.bib and bibliofile.bib.
%% BibTeX documentation can be obtained at:
%% http://www.ctan.org/tex-archive/biblio/bibtex/contrib/doc/
%%%%%%

% %% Or you use manual references (pay attention to consistency and the
% %% formatting style!):
% \begin{thebibliography}{9}

% \bibitem{Laport:LaTeX}
% L.~Lamport,
%   \emph{\LaTeX: A Document Preparation System,} 
%   Addison-Wesley, Reading, Massachusetts, USA, 2nd~ed., 1994. 

% \bibitem{GMS:LaTeXComp}
% F.~Mittelbach, M,~Goossens, J.~Braams, D.~Carlisle, and
% C.~Rowley, \emph{The {\LaTeX} Companion,} Addison-Wesley,
% Reading, Massachusetts, USA, 2nd~ed., 2004.

% \bibitem{oetiker_latex}
% T.~Oetiker, H.~Partl, I.~Hyna, and E.~Schlegl, \emph{The Not So Short
%   Introduction to {\LaTeX2e}}, version 5.06, Jun.~20, 2016. [Online].
%   Available: \url{https://tobi.oetiker.ch/lshort/}

% \bibitem{typesetmoser}
% S.~M. Moser, \emph{How to Typeset Equations in {\LaTeX}}, version 4.6,
%   Sep. 29, 2017. [Online]. Available:
%   \url{http://moser-isi.ethz.ch/manuals.html#eqlatex}

% \bibitem{IEEE:pdfsettings}
% IEEE, \emph{Preparing Conference Content for the IEEE Xplore Digital
%   Library.} [Online]. Available:
%   \url{http://www.ieee.org/conferences_events/conferences/organizers/pubs/preparing_content.html}

% \bibitem{IEEE:AuthorToolbox}
% IEEE, \emph{Author Digital Toolbox.} [Online.] Available:
%   \url{http://www.ieee.org/publications_standards/publications/authors/authors_journals.html}

% \end{thebibliography}

\end{document}